\documentclass[sigconf]{acmart}
\AtBeginDocument{%
  }

\definecolor{MyGreen}{HTML}{3FA796}
\copyrightyear{2026}
\acmYear{2026}
\setcopyright{cc}
\setcctype{by}
\acmConference[WWW '26]{Proceedings of the ACM Web Conference 2026}{April 13--17, 2026}{Dubai, United Arab Emirates}
\acmBooktitle{Proceedings of the ACM Web Conference 2026 (WWW '26), April 13--17, 2026, Dubai, United Arab Emirates}
\acmPrice{}
\acmDOI{10.1145/3774904.3792912}
\acmISBN{979-8-4007-2307-0/2026/04}
\usepackage[dvipsnames]{xcolor}
\usepackage{siunitx, colortbl}
\usepackage{booktabs} % for \toprule, \midrule, \bottomrule
\usepackage{multirow}

\usepackage{amssymb}  % for checkmark
\usepackage[table]{xcolor} % for cell colors
\usepackage{threeparttable} % optional for cleaner table environments
\usepackage{array}
\usepackage{hhline}
\usepackage{natbib}
\usepackage{graphicx}
\usepackage{diagbox}
\usepackage[many]{tcolorbox}

\tcbset{contrib/.style={
    colback=gray!15,
    colframe=gray!40,
    boxrule=0pt,
    arc=2pt,
    left=0pt,right=0pt,top=0pt,bottom=0pt,
}}

\usepackage[framemethod=tikz]{mdframed}

\newmdenv[
  linewidth=0.8pt,
  linecolor=black,
  backgroundcolor=MyGreen!15, % ← 浅灰背景（可改成 black!3~10 调整深浅）
  innerleftmargin=6pt, innerrightmargin=6pt,
  innertopmargin=6pt, innerbottommargin=6pt,
  skipabove=6pt, skipbelow=6pt
]{mybox}

\newcommand{\cA}{\mathcal{A}}
\newcommand{\cC}{\mathcal{C}}
\newcommand{\cX}{\mathcal{X}}

\newcommand{\cP}{\mathcal{P}}
\newcommand{\cL}{\mathcal{L}}
\newcommand{\cS}{\mathcal{S}}
\newcommand{\cG}{\mathcal{G}}

\newcommand{\bx}{\mathbf{x}}

\newcommand{\bz}{\mathbf{z}}
\newcommand{\bt}{\mathbf{t}}

\begin{document}

\title{Breaking Semantic-Aware Watermarks via LLM-Guided Coherence-Preserving Semantic Injection}

\author{Zheng Gao}
\email{zheng.gao1@unsw.edu.au}
\orcid{0009-0004-0330-3493}
\affiliation{%
  \institution{University of New South Wales}
  \city{Sydney}
  \state{New South Wales}
  \country{Australia}
}
\author{Xiaoyu Li}
\email{xiaoyu.li2@student.unsw.edu.au}
\orcid{0009-0002-0078-3829}
\affiliation{%
  \institution{University of New South Wales}
  \city{Sydney}
  \state{New South Wales}
  \country{Australia}
}
\author{Zhicheng Bao}
\email{zhicheng.bao@unsw.edu.au}
\orcid{0000-0003-2692-6752}
\affiliation{%
  \institution{University of New South Wales}
  \city{Sydney}
  \state{New South Wales}
  \country{Australia}
}

\author{Xiaoyan Feng}
\email{xiaoyan.feng@griffithuni.edu.au}
\orcid{0009-0007-6296-2431}
\affiliation{%
  \institution{Griffith University}
  \city{Brisbane}
  \state{Queensland}
  \country{Australia}
}

\author{Jiaojiao Jiang}
\email{jiaojiao.jiang@unsw.edu.au}
\orcid{0000-0001-7307-8114}
\affiliation{%
  \institution{University of New South Wales}
  \city{Sydney}
  \state{New South Wales}
  \country{Australia}
}

\begin{abstract}
Generative images have proliferated on Web platforms in social media and online copyright distribution scenarios, and semantic watermarking has increasingly been integrated into diffusion models to support reliable provenance tracking and forgery prevention for web content. Traditional noise-layer-based watermarking, however, remains vulnerable to inversion attacks that can recover embedded signals. To mitigate this, recent content-aware semantic watermarking schemes bind watermark signals to high-level image semantics, constraining local edits that would otherwise disrupt global coherence. Yet, large language models (LLMs) possess structured reasoning capabilities that enable targeted exploration of semantic spaces, allowing locally fine-grained but globally coherent semantic alterations that invalidate such bindings.
To expose this overlooked vulnerability, we introduce a Coherence-Preserving Semantic Injection (CSI) attack that leverages LLM-guided semantic manipulation under embedding-space similarity constraints. This alignment enforces visual-semantic consistency while selectively perturbing watermark-relevant semantics, ultimately inducing detector misclassification. Extensive empirical results show that CSI consistently outperforms prevailing attack baselines against content-aware semantic watermarking, revealing a fundamental security weakness of current semantic watermark designs when confronted with LLM-driven semantic perturbations.

\end{abstract}

\begin{CCSXML}
<ccs2012>
   <concept>
       <concept_id>10010147.10010257.10010321</concept_id>
       <concept_desc>Computing methodologies~Machine learning algorithms</concept_desc>
       <concept_significance>500</concept_significance>
       </concept>
   <concept>
       <concept_id>10002978.10002991.10002996</concept_id>
       <concept_desc>Security and privacy~Digital rights management</concept_desc>
       <concept_significance>500</concept_significance>
       </concept>
 </ccs2012>
\end{CCSXML}

\ccsdesc[500]{Computing methodologies~Machine learning algorithms}
\ccsdesc[500]{Security and privacy~Digital rights management}
\keywords{Watermark, Large Language Model, Cybersecurity, Diffusion Model}


\maketitle

\section{Introduction}

\begin{figure*}[!ht]
    \centering
    \includegraphics[width=1\linewidth]{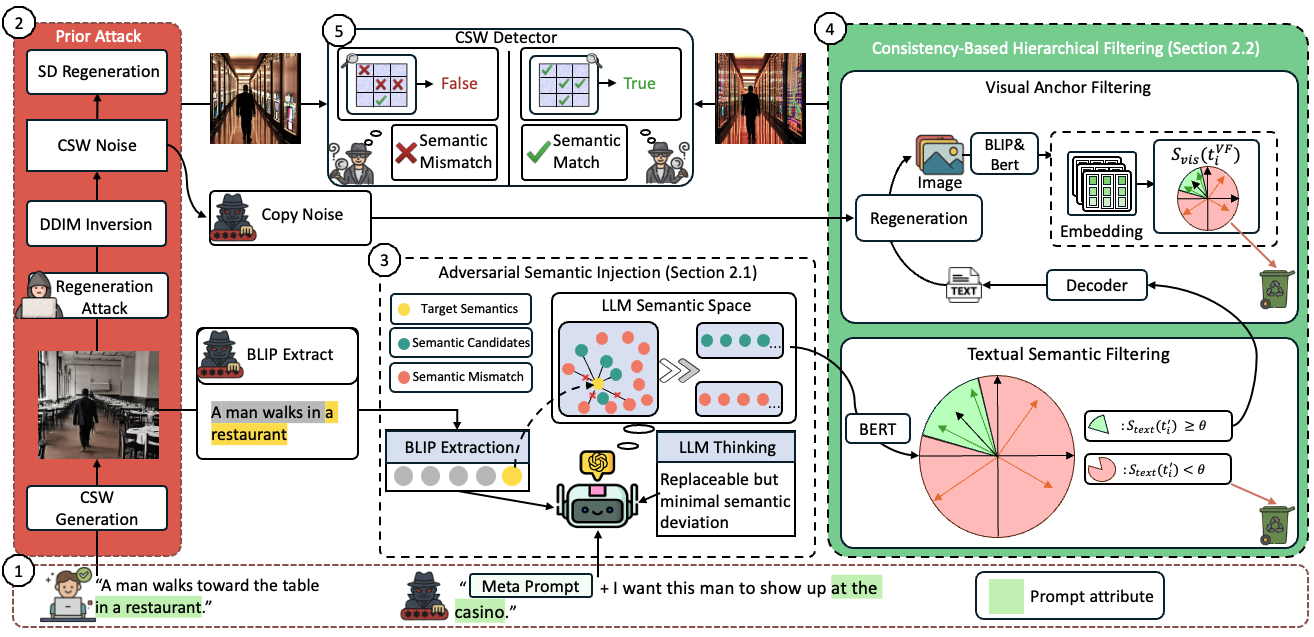}
   \caption{The overall workflow of the proposed Coherence-Preserving Semantic Injection (CSI) attack.}

    \label{fig:model}
\end{figure*}

With the widespread adoption of diffusion-based generative models across the Web ecosystem, AI-generated images have become increasingly pervasive on social media platforms and ever harder to distinguish from real photographs~\cite{yang2023diffusionSurvey}, making content authenticity and copyright traceability increasingly crucial. Pixel-level watermarking provides a direct mechanism for provenance, but it faces a fundamental trade-off between visual fidelity and robustness, as watermarks are easily degraded by compression, filtering, and other transformations during online dissemination. To mitigate this, \emph{semantic watermarking} embeds information into the diffusion noise rather than visible pixels, e.g., Tree-Ring Watermark, Gaussian Shading, and WIND, which encode signals in the initial (or latent) noise following the model’s generative prior and recover them via approximate inversion of the diffusion process~\cite{wen2023treering-article,Yang24GaussianShading-long,arabi2024hiddennoise-article}. By embedding the watermark in semantic noise, these methods typically preserve perceptual quality while achieving strong robustness to common perturbations, making them promising candidates.

However, these semantic watermarking methods share a key vulnerability: they embed watermark signals mainly in the initial latent noise with weak semantic alignment, which we refer to as Content-independent Semantic Watermarks (CIW). Exploiting this weakness, Müller et al.\ show that a single watermarked image suffices to recover the watermark and generate arbitrary content with forged marks~\cite{Muller2025BlackBox}. To mitigate this, Arabi et al.\ propose Semantic-Aware Image Watermarking (SEAL)~\cite{Arabi25SEAL-long}, a Content-aware Semantic Watermark (CSW) scheme that conditions verification on image semantics rather than noise alone and tightly couples inverted noise with visual content, forcing attackers to preserve semantic consistency and making effective forgeries substantially harder.

In other words, attacking such watermarks essentially requires solving a multi-constrained semantic optimization problem in a discrete prompt space: the attacker must preserve the coherence between the primary visual semantics and the noise semantics while injecting adversarial attributes to alter local semantics. 
Coincidentally, recent studies show that LLMs can not only automate combinatorial optimization tasks requiring strict constraint definitions, but also heuristically search for optimal semantic solutions in discrete prompt spaces~\cite{yang2023large,zhang2025or}. This emerging capability indicates that the original security assumptions are fundamentally flawed. 

Based on the above insight, we propose \tcbox[contrib,on line]{\textsc{\textbf{(contribution i)}}:} \textbf{a Coherence-Preserving Semantic Injection (CSI)} attack, which uses Adversarial Semantic Injection via Semantically Coherent Manipulations (ASI) combined with a Consistency-Based Hierarchical Filtering (CHF) mechanism to ensure successful semantic alteration while maintaining positive watermark detection. 

To the best of our knowledge, CSI is the first systematic attack against CSW such as SEAL. \tcbox[contrib,on line]{\textsc{\textbf{(contribution ii)}}:} \textbf{Our experiments} reveal that even state-of-the-art CSW cannot withstand CSI, underscoring a critical security gap and highlighting the urgent need for more robust, hierarchical watermarking mechanisms capable of defending against semantic-level adversarial attacks.

\section{Method: Coherence-Preserving Semantic Injection}

In this section, we introduce our method: \emph{Coherence-Preserving Semantic Injection} (CSI), as illustrated in Figure~\ref{fig:model}. We first define some useful notations.
Let $\bx_0$ be the semantic-aware watermarked image, and $\bt_0 = \mathrm{BLIP}(\bx_0)$ be the associated semantic with it generated by BLIP model. We use $\{\epsilon_t\}_{t = 1}^T$ to denote the noise schedule of the stable diffusion. Let $\cG$ denote the set of global semantic anchors (e.g., subjects/main objects), and $\cA$ the local attribute to manipulate. We denote by $\Pi_{\mathcal G}$ the mask operator that keeps only anchor tokens, and by $a^\star$ the target attribute specified by the attack intent $I$. Let $\phi_{\mathrm{img}}$, $\phi_{\mathrm{text}}$, and $\phi_{\mathrm{noise}}$ be the normalized image, text, and noise encoders, respectively. Thus $\langle\cdot,\cdot\rangle$ is the cosine similarity in the unit sphere.

\subsection{Adversarial Semantic Injection via Semantically Coherent Manipulations (ASI)}

{\bf Conceptual objective.}
We seek a modified prompt $\bt'$ that (i) preserves global anchors and (ii) injects the target attribute, while (iii) allowing diffusion regeneration that matches CSW noise semantics:
\begin{align}
\bt' \in~ \arg\min_{\bt}~
&\lambda_{\mathrm{anc}}\,\cL_{\mathrm{anc}}(\bt,\bt_0)
-\lambda_{\mathrm{attr}}\,\cS_{\mathrm{attr}}(\bt,a^\star) \notag \\
\text{s.t.}\quad
& \bx'=\mathrm{SD\_regen}(\bz_T,\bt,\{\epsilon_t\}_{t=1}^{T})  \label{eq:obj}
\\
& s_{\mathrm{csw}}(\bx', \{\epsilon_t\}_{t=1}^T) \ge \tau_{\mathrm{csw}} \notag
\end{align}
where the anchor-preservation loss $\cL_{\mathrm{anc}}$ penalizes deviation on anchor tokens, the attribute-injection score $\cS_{\mathrm{attr}}$ rewards the presence of the target attribute, and the constraints are defined below.

{\bf Regeneration with copied noise.}
We use DDIM inversion and noise copying to save the watermark-consistent noise semantics:
\begin{equation}
\bz_T=\mathrm{DDIM\_inv}(\bx_0),\qquad
\{\epsilon_t\}_{t=1}^T=\mathrm{CSW\_Noise}(\bx_0).
\end{equation}
Given any candidate prompt $\bt$, the image $\bx'$ is regenerated as:
\begin{equation}
\bx'=\mathrm{SD\_regen}\left(\bz_T,\bt,\{\epsilon_t\}_{t=1}^{T}\right).
\end{equation}
We define the CSW score to quantify the semantic alignment between the image and the noise:
\begin{align}
    s_{\mathrm{csw}}(\bx,\{\epsilon_t\})
&=\left\langle \phi_{\mathrm{img}}(\bx), \phi_{\mathrm{noise}}(\{\epsilon_t\}_{t=1}^T) \right\rangle.
\end{align}

\definecolor{MyGreen}{RGB}{180, 232, 200}
\definecolor{HeaderGray}{gray}{0.9}

\begin{table}[t]
\centering
\small
\renewcommand{\arraystretch}{1.2}
\begin{threeparttable}
\caption{Comparison of different attacks under each detector (ASR \%↑).}
\label{tab:detector_comparison}
\begin{tabular}{l|cc|>{\columncolor{MyGreen!55}}c}
\hline
\rowcolor{HeaderGray}
\textbf{Detector (ASR\tnote{a}\ \ \% ↑)} & \textbf{LFA~\cite{jain2025forging-article}} & \textbf{RPM~\cite{Arabi25SEAL-long}} & \textbf{CSI (Ours)}\\
\hline
\multicolumn{4}{c}{\textit{Content-independent semantic watermarks}} \\
\hline
Gaussian Shading~\cite{Yang24GaussianShading-long} & 100 & 100 & \textbf{100}  \\
Tree-Ring~\cite{wen2023treering-article} & 93.81 & 100 & \textbf{100}  \\
WIND~\cite{arabi2024hiddennoise-article} & 100 & 100 & \textbf{100} \\
\hline
\multicolumn{4}{c}{\textit{Content-aware semantic watermark}} \\
\hline
SEAL~\cite{Arabi25SEAL-long} & 0 & 7 & \textbf{81}\\
\hline
\textbf{Mean ↑} & 73.45 & 76.75 & \textbf{95.25} \\
\hline
\end{tabular}
\begin{tablenotes}[flushleft]\footnotesize
\item[a] ASR = proportion of tampered images still detected as “watermark present” under the threshold.
\end{tablenotes}
\end{threeparttable}
\end{table}

{\bf Approximate optimization by LLM.}
Since directly optimizing the conceptual objective~\eqref{eq:obj} over discrete tokens under complex constraints is unstable and intractable, motivated by \cite{yang2023large,zhang2025or}, we adopt an optimization-by-prompting approach. We treat the LLM as a black-box proposer and craft a meta-prompt that specifies the objective and constraints in natural language. Conditioned on $(\bx_0, \bt_0, \cG, I)$, the LLM generates a batch of semantically coherent prompt candidates, which we then filter to enforce anchor preservation and attribute injection. Hence it prevents prompt collapse while preserving grammaticality. During regeneration, we reuse the copied CSW noise so that any change in detector acceptance can be attributed to semantic edits rather than stochasticity.

\begin{mybox}
    {\textbf{Meta Prompt:} You are a prompt generator for a watermark evasion study. Each prompt must keep the same main subject \textbf{[Name]} and allow only minor semantic changes \textbf{[Modification Target]}, ensuring the overall meaning stays consistent.} 
\end{mybox}

\subsection{Consistency-Based Hierarchical Filtering (CHF)}

The LLM optimizer yields a pool $\cP$ of minimally edited prompts. 
We now filter $\cP$ by progressively stronger tests that are (i) cheap and text-only, then (ii) visual alignment and CSW-aware. 
Define the following acceptance sets for thresholds $\tau =(\tau_{\mathrm{text}},\tau_{\mathrm{vis}},\tau_{\mathrm{csw}})$:

{\bf Textual Semantic Filtering.} We first remove candidates that drift from the global anchors using a text-only check:
\begin{align}
    s_{\mathrm{text}}(\bt'_i) = \left\langle \phi_{\mathrm{text}}(\Pi_{\mathcal G}\bt_0),\,\phi_{\mathrm{text}}(\Pi_{\mathcal G}\bt'_i) \right\rangle.
\end{align}
Candidates that preserve anchors form
\begin{align}
\cC = \{ \bt'_i \in\cP : s_\mathrm{text}(\bt'_i) \geq \tau_{\mathrm{text}} \}.
\end{align}

{\bf Visual Anchor Filtering.} For each $\bt'_i\in\cC$, we regenerate an image $\bx'_i$ with the copied CSW noise to neutralize sampling randomness, and use BLIP to get its caption $\bt^{\mathrm{VF}}_i = \mathrm{BLIP}(\bx'_i).$
We then compute the visual-anchor alignment, which reflects whether the anchors preserved in the image:
\begin{align}
    s_{\mathrm{vis}}(\bt^\mathrm{VF}_i)=\left\langle \phi_{\mathrm{text}}(\Pi_\cG\bt_0), \phi_{\mathrm{text}}(\Pi_\cG\bt^{\mathrm{VF}}_i) \right\rangle 
\end{align}
Finally, we impose CSW semantic matching between the regenerated image and the copied noise schedule via the discrepancy:
\begin{align}
\Delta_{\mathrm{csw}}(\bx'_i) = 
1-\left\langle \phi_{\mathrm{img}}(\bx_i'),\phi_{\mathrm{noise}}(\{\epsilon_t\}_{t=1}^T)\right\rangle.
\end{align}
The final set of attack images is
\begin{align}
    \cX_{\mathrm{attack}} = \left\{ \bx'_i: \bt'_i \in \cC, s_\mathrm{vis}(\bt'_i) \geq \tau_{\mathrm{vis}}, \Delta_{\mathrm{csw}}(\bx'_i) \leq \tau_{\mathrm{csw}} \right\}.
\end{align}

\section{Experiments}
In this experiment, We employ the Stable Diffusion V2 model as the generation model. The evaluation prompts are sourced from the publicly Stable-Diffusion-Prompts dataset~\cite{Santana2024StableDiffusionPrompts}, and we use GPT-4o-mini as the large language model.

\textbf{Attack baseline}: We compare our attack method with the semantic attacks prevalent in literature, including Regeneration with the Private Model (RPM)~\cite{Arabi25SEAL-long} and Latent Forgery Attack (LFA)~\cite{jain2025forging-article}.

\textbf{Semantic watermark baseline}: We select four semantic watermark techniques for defense, including Semantic Aware Image Watermark (SEAL)~\cite{Arabi25SEAL-long}, Gaussian Shading Watermark (GSW)~\cite{Yang24GaussianShading-long}, Tree-Ring Watermark (TRW)~\cite{wen2023treering-article} and WIND~\cite{arabi2024hiddennoise-article}.

\subsection{Comparative Attack Effectiveness}

To evaluate the effectiveness of our attack, we conduct comparative experiments involving multiple semantic attacks under various semantic watermark defenses. We use the attack success rate (ASR) $\uparrow$, defined as the proportion of tampered images still detected as watermark present under the threshold, as the evaluation metric. The experimental results are shown in Table~\ref{tab:detector_comparison}.

From Table~\ref{tab:detector_comparison}, we observe that our method, along with the two baseline attacks, achieves nearly 100\% ASR against the GSW, TRW, and WIND semantic watermarks. This suggests that these watermarking schemes are highly vulnerable to semantic attacks and can be easily compromised. However, under the protection of the content-aware semantic watermarking method SEAL, RPM and LFA fail almost entirely, achieving only 7\% and 0\% ASR, respectively. In contrast, our method still achieves an impressive 81\% ASR, clearly demonstrating its robustness and effectiveness in bypassing even advanced semantic watermark defenses.

\begin{figure*}
    \centering
    \includegraphics[width=0.98\linewidth]{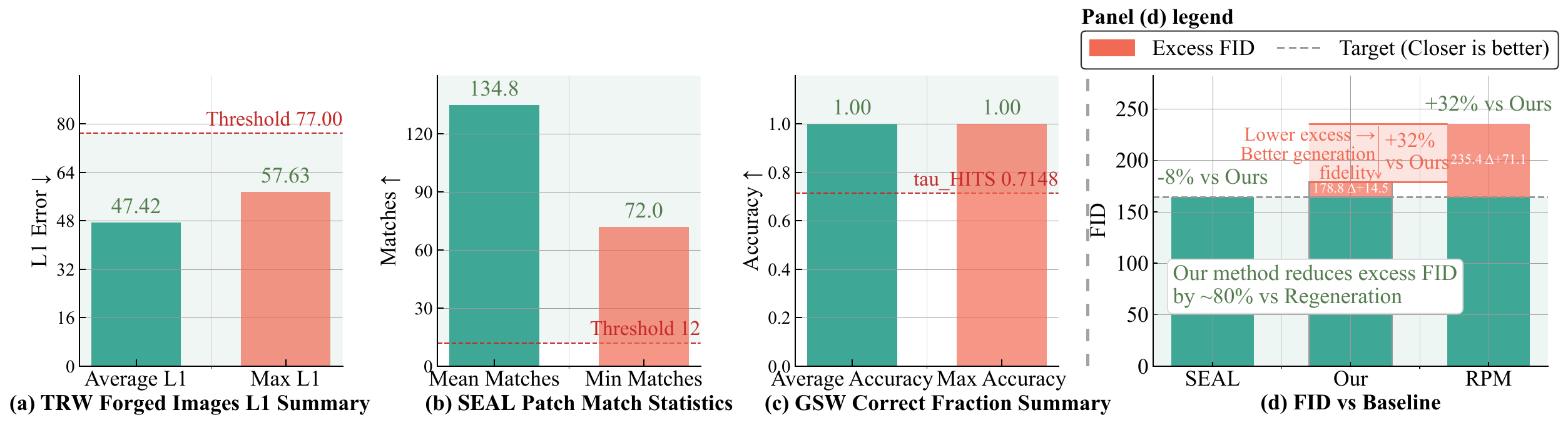}
    \caption{Further Analysis of Detection Metrics}
    \label{fig:experiment}
\end{figure*}
\subsection{Attack Robustness Across Different Watermarking Schemes}

To quantify the vulnerability of existing semantic watermark detectors to our attack, we measure, for each scheme, its core detection statistic and compare it against the official decision threshold. We further report the corresponding evasion success rate, numerical safety margin, and worst-case deviation, thereby establishing a consistent and verifiable basis for evaluating attack effectiveness across different watermarking mechanisms. We detail the evaluation of four representative schemes below:

\begin{enumerate}
    \item \textbf{TRW}, which detects forgeries by measuring the L1 distance between reconstructed and reference noise patterns. The attacked images yields an average distance of only 47.42 in our attack (Figure~\ref{fig:experiment} a), far below its detection threshold. Even the maximum value reaches merely 57.63, maintaining a clear safety margin of 19.37.
    \item \textbf{SEAL}, which enforces semantic consistency by comparing the number of matching patches between the inverted noise $\mathbf{z}^{\text{inv}}$ and the regenerated reference $\tilde{\mathbf{z}}$. Our method attains an average match count of 134.8 (Figure~\ref{fig:experiment} b), with even the lowest case reaching 72, far exceeding the threshold of 12.
    \item \textbf{GSW}, which verifies authenticity by decoding a $K$-bit watermark sequence $s’$ from the inverted initial noise and comparing it with the original $s$. Our method achieves a perfect matching score of 1.00 (Figure~\ref{fig:experiment} c), well above the detection threshold of 0.71.
    \item \textbf{WINE}, which authenticates images by matching the reconstructed noise against a secret noise pattern. Our method consistently produces exact matches across all attacked samples, indicating complete evasion of detection.

\end{enumerate}

Overall, the results confirm that our attack consistently lowers the measured detection scores across all other schemes, demonstrating robust evasion capability under diverse detector.

\subsection{Attack Effectiveness on Content-aware Watermarking}

To evaluate our attack against content-aware semantic watermarking, we examine whether it disrupts the core detection criterion: semantic consistency between the image and its prompt. This consistency is defined as the similarity of their semantic representations. At the set level, this reflects distribution alignment in high-level semantic feature space: if the tampered set matches the original distribution, global semantics are preserved, potentially bypassing detection. We quantify this using Fréchet Inception Distance (FID)~\cite{heusel2018ganstrainedtimescaleupdate}, which compares the means and covariances of image sets to capture semantic distribution differences.

Based on this, we conduct semantic consistency experiments to assess the impact of LLM-based semantic constraints on preserving semantics after tampering. We compare our method (Ours, LLM-constrained) against two baselines: RPM (unconstrained regeneration) and SEAL (watermarked and unaltered). The reported FID values are computed between the generated images and the original image set. The results are illustrated in Figure ~\ref{fig:experiment}(d).

Results show that RPM exhibits an FID of 235.40, indicating substantial semantic drift and confirming that unconstrained regeneration severely disrupts the original semantic distribution. In contrast, Ours achieves an FID of 178.75, an absolute reduction of 56.65 (↓24.1\%) compared to RPM, and approaches SEAL’s 164.27. Importantly, in this setting, the goal is not to minimize FID arbitrarily but to align as closely as possible with SEAL, as this reflects successful semantic preservation while performing the attack. It demonstrates that LLM semantic constraints significantly mitigate semantic drift and enable tampering without deviating from the watermark’s content-aware semantic structure, validating their critical role in bypassing content-aware semantic watermark detection.

\section{Conclusion}
Our findings indicate that even content-aware semantic watermarking schemes such as SEAL remain vulnerable to LLM-guided semantic attacks. To validate this vulnerability, we propose the CSI framework, demonstrating that LLMs can systematically compromise watermark integrity within the discrete semantic space. These findings highlight a critical security gap and the need for future watermark designs to defend against semantic-level attacks.

\bibliographystyle{ACM-Reference-Format}
\bibliography{ref}

\end{document}